# Ranking and Rejecting of Pre-Trained Deep Neural Networks in Transfer Learning based on Separation Index

Mostafa Kalhor [1], Ahmad Kalhor *,[2] Mehdi Rahmani [1]

**Abstract**— Automated ranking of pre-trained Deep Neural Networks (DNNs) reduces the required time for selecting optimal pre-trained DNN and boost the classification performance in transfer learning. In this paper, we introduce a novel algorithm to rank pre-trained DNNs by applying a straightforward distance-based complexity measure named Separation Index (SI) to the target dataset. For this purpose, at first, a background about the SI is given and then the automated ranking algorithm is explained. In this algorithm, the SI is computed for the target dataset which passes from the feature extracting parts of pre-trained DNNs. Then, by descending sort of the computed SIs, the pre-trained DNNs are ranked, easily. In this ranking method, the best DNN makes maximum SI on the target dataset and a few pre-trained DNNs may be rejected in the case of their sufficiently low computed SIs. The efficiency of the proposed algorithm is evaluated by using three challenging datasets including Linnaeus 5, Breast Cancer Images, and COVID-CT. For the two first case studies, the results of the proposed algorithm exactly match with the ranking of the trained DNNs by the accuracy on the target dataset. For the third case study, despite using different preprocessing on the target data, the ranking of the algorithm has a high correlation with the ranking resulted from classification accuracy.

[1] Department of Electrical Engineering Imam-Khomeini International University Qazvin, Iran
[2] Corresponding Author, School of Electrical and Computer Engineering, University of Tehran
  E-mail address: akalhor@ut.ac.ir (A. Kalhor)

# 1. Introduction

During the past decade, deep neural networks have been progressing rapidly compared to traditional machine learning algorithms [1] in terms of speech recognition, natural language processing, compute vision [2,3,4], especially in the classification of large-scale dataset like ImageNet [5]. In the classification problems, one of the major benefits of DNNs is that they can discover automatically common features of the dataset. However, there are some main limitations in the use of DNNs [6], which are highlights as follows:

- A powerful graphics processing unit (GPU) is needed for the learning process.
- DNNs need a large number of labeled training data to have better classification accuracy.

In order to address these limitations, the best solution would be the concept of transfer learning. In transfer learning, the technical goal is to extract the knowledge from the source task for applying it to a target task. The first motivation for transfer learning was presented in a NIPS-95 workshop on "Learning to learn", which gained significant research interest at that point [7]. At first, the important question was which part of knowledge can be transferred between tasks.

To answer this question, transfer learning is categorized into four major classes including transferring knowledge of instances [8], transferring knowledge of feature representations [9], transferring knowledge of Parameters [10] and transferring relational knowledge [11]. In transferring knowledge of instances, the source task is re-weighted to be used for the target task [12]. The feature-representation-transfer concept tries to discover appropriate features of the source task to be applied for the decrease of classification or regression model error [13]. The main goal of parameter-based transfer learning is to evaluate the transformed parameters between tasks [14]. The relational-knowledge-transfer approach attempts to discover the relationship among data from source task to target task [15].

In classification problems using transfer learning, the weights of the feature extracting part of DNNs, which are learned from the source dataset, are reused as a starting point for the classification of the target dataset and the partitioning part can be redesigned and learned through a limited target dataset. It is observed that transfer learning has improved the classification accuracy at an unprecedented rate in many fields compared to other traditional methods [16, 17].

Despite the high performance of this method, there is still a challenging problem when the user concerns about choosing optimal pre-trained DNNs for a target dataset, which has become more challenging due to the large number of datasets and pre-trained DNNs. As it is clear, the selection of pre-trained DNN has a great impact on the improvement of classification accuracy. At first, some approaches were applied for manually choosing pre-trained DNNs in transfer learning [18, 19], which have two major drawbacks:

- It needs to evaluate manually a large number of pre-trained DNNs, which could be a time-consuming process.
- It is not reliable since the selected pre-trained DNN would lead to worse performance compared to others.

As a result, these drawbacks have derived many researchers to design and develop various frameworks to predict automatically the performance of candidate pre-trained DNNs for a given target dataset. In [20], it is aimed to investigate the source-target relationship using the computing of transferability index (entropy function) to design an automatic framework, which can rank pre-trained DNNs in a zero-shot manner. It is shown that there is a high correlation between the computed transferability index and classification accuracies obtained after transfer learning. In [21], the Representation Similarity Analysis (RSA) is used to compute a similarity score between tasks using calculating correlations between different pre-trained DNNs. The efficiency of this approach is shown by evaluating the relationship between the performance of transfer learning on Taskonomy [22] tasks and a new task. In this work, the results indicate that models, which are trained on tasks with more similarity scores, show better transfer learning performance. In [23], Duality diagram similarity (DDS) is applied to compare the relationship between features of pre-trained DNNs with new task features to select the optimal model initialization for learning a new task. The task feature is obtained by doing a feedforward pass through a model trained on the source task.

In some previous works [24, 25], it has been observed that some predictors are applied to select suitable pre-trained DNN for the classification of a target dataset. One of the most considerable accuracy predictors is Train-less Accuracy Predictor for Architecture Search (TAPAS) [26], which works based on source-target Dataset Classification Difficulty (DCN) Estimation. This framework predicts the performance of the candidate network, which consists of two stacked Long Short-Term Memory (LSTM) Networks. In this approach, TAP must be trained by a lot of source datasets through the Lifelong Database of Experiments (LDE), which are similar to the target dataset in terms of dataset complexity. The trained TAP takes only Dataset Characterization Number of target dataset and candidate network structure as input to forecasts peak accuracy of candidate network without any new training experiment. In TAP, Dataset Classification Difficulty Estimation can be computed through different indices [25] including silhouette score, k-means clustering, Fréchet inception distance-based score, and probe nets. Due to the efficient performance of probe nets, this index is used as an estimator of DCN in [25]. In this work, TAPAS applies a genetic algorithm to find the optimal classifier model.

In [27], the concept of Neural Architecture Search (NAS) is investigated for finding well-performing DNNs in its search space that gain high predictive performance on a new target dataset. At first, this method was faced with some limitations because of the large search space. In [28], Efficient Neural Architecture Search (ENAS) is proposed to reduce the search space via a directed acyclic graph. In this framework, there is an LSTM neural network to control and make a decision about the kind of candidate network layers. This concept is used in [24], where the main goal is to select optimal pre-trained DNNs based on the DCN of the source-target dataset. In this method, the framework takes a target dataset and calculates its DCN to select a subset of data (source dataset) from the LDE, which has the nearest DCN to the target dataset. Then, the ENAS controller is trained on the selected data to find the optimal pre-trained DNN and predict its accuracy for the given dataset.

Although the previous proposed methods provide some advantages in the automated ranking of the pre-trained DNNs, each one may suffer from some limitations and bottlenecks:

- In some frameworks, using source data is essential [22, 24, 25]. This avoids evaluating some pre-trained DNNs, which there is no access to their source datasets.
- In order to rank the pre-trained DNNs, some methods need to run time-consuming training processes. This increases the computation load, particularly when the number of pre-trained DNNs is high, [24, 25].
- In [20], due to applying complex statistical analysis on the training dataset, the computation load can be increased.
- Many proposed methods only rank the pre-trained DNNs and do not give any criterion to reject some pre-trained DNNs in transfer learning [20, 21 22, 23, 24, 25].

The above-mentioned limitations and bottlenecks pose a question that whether it is possible to rank and reject pre-trained DNNs with a simple and fast evaluation index and without considering the connection between the source-target dataset and extra training processes. In this work, a novel algorithm is proposed to overcome the aforementioned problems which can be used to rank and also reject the considered pre-trained DNNs for classification of a target dataset. In the proposed algorithm, only the target dataset complexity of the last layer of pre-trained DNNs are computed and compared. For this purpose, the Separation Index (SI), which is a straightforward distance-based geometric factor, is applied. The major advantages of the proposed algorithm are summarized as follows:

- Due to using a simple geometric-based complexity index, applying the proposed algorithm is straightforward in the automated ranking of pre-trained DNNs.
- The proposed algorithm does not require an understanding of the relationship between the source-target dataset.
- There is not any training process in this proposed algorithm to rank pre-trained DNNs.
- This proposed algorithm can reject the pre-trained DNNs for a specific target dataset, as well as ranking them.
- The sensitivity of the algorithm to the number of training data points is low. The resulted ranking for pre-trained DNNs does not change by a limited reduction of the dataset.

The organization of the paper is as follows: In section 2, the concept of separation index and the proposed algorithm are discussed. To show the efficiency of the proposed algorithms, some case studies are presented in section 3. Finally, in section 4, conclusion remarks are given.

## 2. Methodology

*2.1. The Concept and Definition of Separation Index*

In supervised classification problems, the analysis of dataset complexity plays a main role in the prediction of the classification accuracy [29], which can be used to support the selection of optimal classifier. Over the last few decades, many papers have applied complexity measures to characterize the dataset and to get better prediction results [30, 31]. Generally, the complexity measures can be mainly divided into six main categories [32], which are summarized as follows:

- Feature-based measures, which their main goal is to discover informative features to classify the classes. Most of these measures try to evaluate each feature independently and are applied for binary classification problems. It is clear that the more discriminative features are characterized through these measures, the more simple the problem will be. The practical measures which belong to this group are the maximum fisher's discriminant ratio, the directional-vector maximum fisher's discriminant ratio [33] and the volume of overlapping region [34].
- Linearity measures, which are applied to determine to what extent the classes are linearly separable. As it is obvious, linearly separable problems are simpler compared to classification problems that need a non-linear hyperplane as a classifier. The known measures in this group are the sum of the error distance by linear programming [35] and the error rate of linear classifier.
- Neighborhood measures, which evaluate the shape of the decision boundary to distinguish the classes overlap using analysis of local neighborhoods of the dataset. These kinds of measures work based on the distance between all pairs of points in the dataset. In this case, there are some important measures including the fraction of borderline points [36], the error rate of the nearest neighbor classifier and the local set average cardinality [37].

- Network measures, which obtain structural information from the dataset. It is vital to show the dataset as a graph to utilize these measures. In this concept, the represented graph store the distances between dataset to model the data relationships. Some measures in this group are the average density of the network and the clustering coefficient [38].

- Dimensionality measures, which are able to give helpful information regarding data sparsity. These measures work based on the dataset dimensionality. It can be seen that the sparse dataset can result in a difficult classification problem. There are known measures in this category including the average number of features per dimension [39] and the average number of PCA dimensions per points [36].

- Class imbalanced measures, which try to measure the proportion of dataset number between classes. In this concept, if the dataset has a high imbalance in the number of per class, the classification problem can be regarded as a complex problem. Some data complexity measures in this group are the entropy of class proportions [36] and the imbalance ratio [40].

As mentioned, many measures could be taken to evaluate the complexity of the dataset for the prediction of classifier performance. The high efficiency of these measures posed an important question that whether it is possible to predict the performance of deep neural networks in the classification problems using computation of dataset complexity. In [41], the complexity of the dataset and feature points are evaluated using a neighborhood measure called Separation Index (SI) [32, 41]. This index is described as a straightforward distance-based measure, which can be applied to analyze dataflow through layers of deep neural networks in the supervised classification problems [32].

To clarify the concept of separation among dataset and feature points with their labels, a top view of revealing hand-made example in a two-dimensional space is shown in Fig.1. From this figure, the dataset with different labels are indicated by green rectangles, black circles, and blue stars. As it is clear, due to the distortions and disturbances, the dataset are in random positions in the vector space. In other words, the dataset with the same label cannot form anymore. For this reason, the different kinds of classifiers like deep neural networks have been designed to categorize the dataset with their labels.

Fig.1 presents two feedforward deep neural networks, which act as a classifier model. These classifiers consist of two key components, including a feature extracting part and a partitioning part. At first, the feature extracting part is responsible for filtering distortions, disturbances and nullities of the given dataset to get a new space with effective features. Then, the partitioning part classifies the feature points by considering boundaries around the local regions, which include dataset with the same labels.

As it is understood from Fig.1 (a), the dataset with the same labels come near together by designing the appropriate feature extracting layers. In other words, the dataset with different labels become separated from each other through layers of deep neural network. In such condition, it is expected that the classifier (a) achieves high classification accuracy. On the other hand, in Fig.1 (b), it is seen that dataset with the non-equal labels remain close to each other. It is evident that the classifier (b) yields low classification accuracy compared to the classifier (a).As observed, the concept of separation can be applied to analyze and predict which classifiers (deep neural networks) are more promising for a specific dataset. For this mean, a practical numerical index called Separation Index can be used.

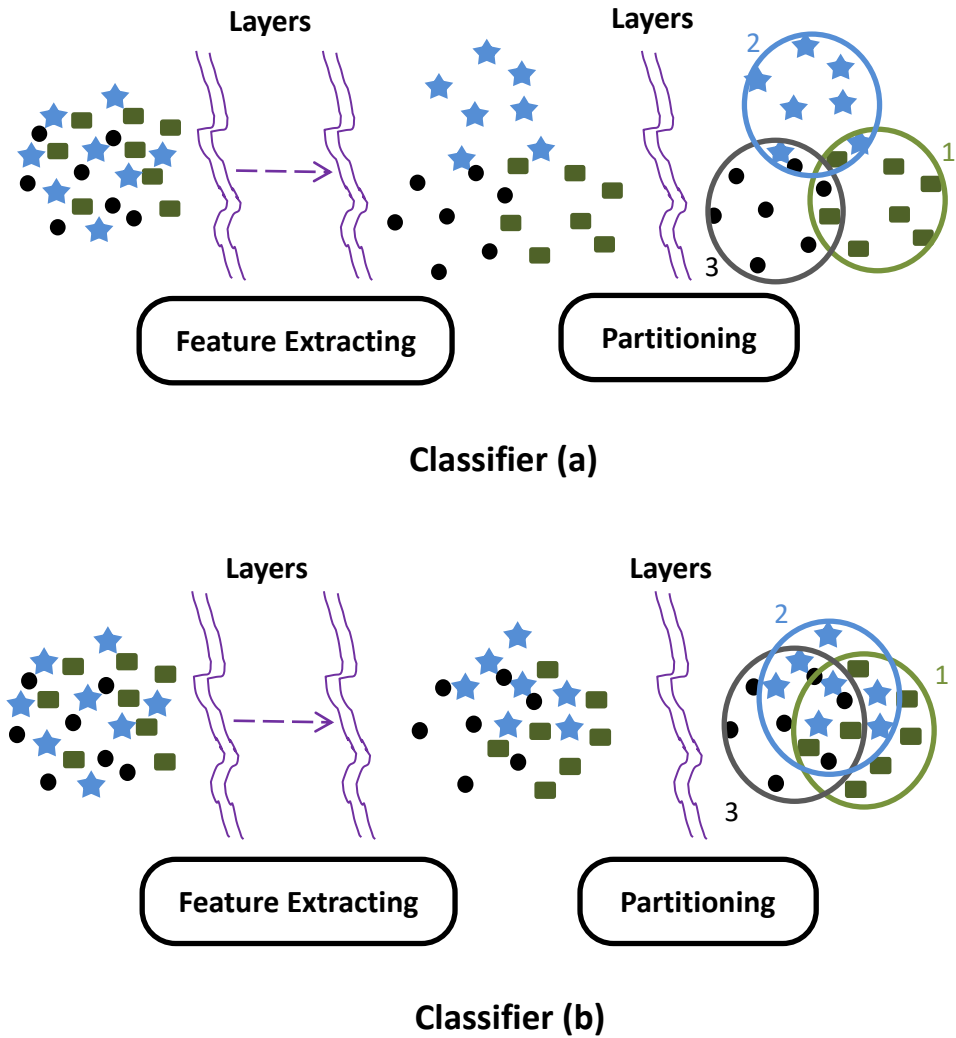

**Fig. 1.** The top view of revealing hand-made example to clarify the concept of separation index. In classifier (a). It is shown that the dataset with same labels come near together through layers of deep neural network leading to the increase of separation index. On the other hand, in classifier (b), it is clear that the dataset with same labels remain still far from each other through layers resulting in the decrease of separation index.

To compute the separation index of a given dataset, first, the Euclidean distance from each point to all points is calculated. Then, the nearest neighbor is chosen for any points to check their labels. As it is clear, if their labels are the same, the separation between data points with various labels is increased. In this way, this measure can help us to characterize the overlap between different classes. To evaluate the dataset and also dataflow through layers of deep neural networks, the separation index can be computed as follows:

$$\text{SI}\left(\{x_L^q\}_{q=1}^Q, \{l^q\}_{q=1}^Q\right) = \frac{1}{Q}\sum_{k=1}^{Q} \varphi\left(l^q - l^{q_{near}^L}\right) \qquad \varphi(v) = \begin{cases} 1 & v = 0 \\ 0 & v \neq 0 \end{cases} \qquad (1)$$

$$q_{near}^L = argmin\|x_L^q - x_L^h\|_h \qquad\qquad h \in \{1, 2, \ldots, Q\} \quad \text{and} \quad h \neq q \quad (2)$$

Where $\{x^q, l^q\}_{q=1}^Q$ and $\varphi(.)$ denotes the dataset with their labels and Kronecker delta function respectively. Also, the $\{x_L^q\}_{q=1}^Q$ presents as dataflow in layer L $\varepsilon$ $\{1,\ldots, N_L\}$. From (1) and (2), it is found the separation index is described as a normalized number, which can be between 0 and 1. The increase of separation index of the dataset in the end layer of deep neural networks (near to one) means that there is enough marginal space between the various classes so that the discriminator hyperplanes can be fitted among them. This increase results in better classification performance.

As observed, in the classification problems, the evaluation of dataset separation through layers of deep neural networks can enable us to predict the network's performance. In this paper, the main goal is to use this approach for ranking and rejecting the pre-trained deep neural networks in transfer learning. To fulfill such goal, an efficient algorithm is proposed to compute and compare the separation index of given dataset in the last layer of available pre-trained DNNs. The details of this proposed algorithm are described in the next section.

*2.2 Proposed Algorithm*

Choosing an appropriate pre-trained deep neural network can affect significantly the classification results in transfer learning. In this section, a novel algorithm is proposed to decrease the time taken for selecting optimal pre-trained DNN. The proposed algorithm applies the aforementioned separation index to ranks and rejects the pre-trained DNNs. At first, the separation index of the given target dataset is computed. Then, the size and input channel number of the target dataset must be changed according to the default of pre-trained DNNs. Next, the pre-processed target dataset are inserted into the pre-trained DNNs. Here, a good idea is to compute the separation index of the target dataset in the end layer of networks. After that, the proposed algorithm sort pre-trained DNNs in descending order based on the computed separation index. Finally, the algorithm rejects pre-trained DNNs, which decrease the separation index of the target dataset through their layers, and rank remained pre-trained DNNs (see Figure.2). According to the aforementioned algorithm, the best pre-trained DNN makes maximum SI on the target dataset. It is expected that this network obtains high performance for a given target dataset. As observed, the proposed algorithm does not utilize the source dataset. In addition, there is no training process in this concept, which could dramatically accelerate the proposed algorithm.

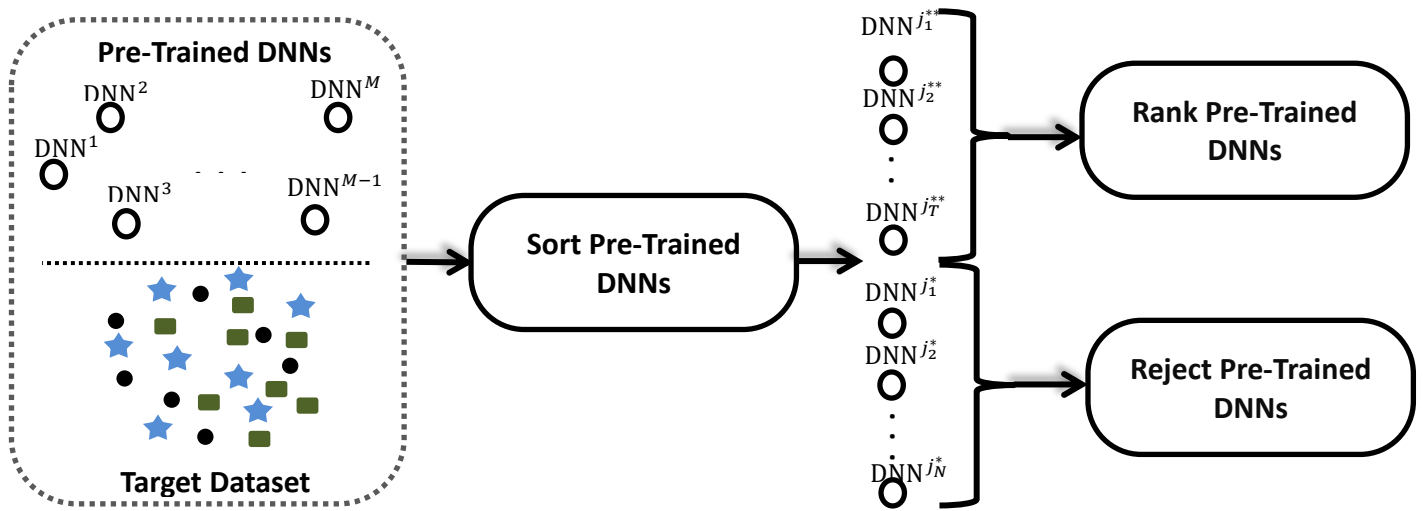

**Fig. 2.** Given a number of pre-trained DNNs, the proposed algorithm ranks and rejects them in a descending order based on the computed separation index of their final layer. The source dataset is not used in this approach.

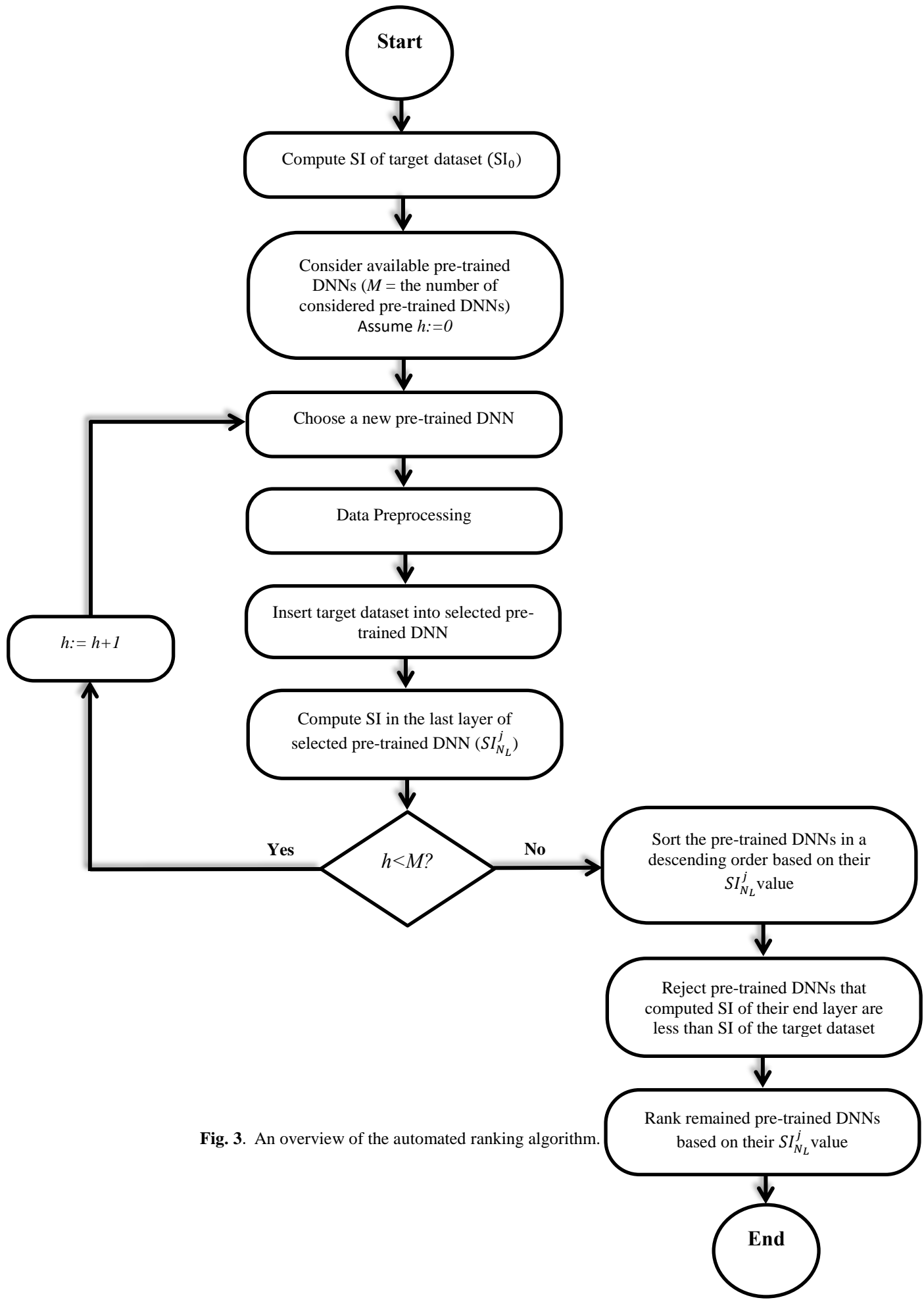

**Fig. 3**. An overview of the automated ranking algorithm.

Respecting to the previous descriptions, an overview of the automated ranking algorithm is presented in Fig. 3. According to the given overview, the proposed algorithm is explained in detail:

- **Step1:** Receive ($\{x^q, l^q\}_{q=1}^Q$) and compute the $SI_0$ in the space x from (1) and (2), where $SI_0$ denotes separation index of target dataset. As mentioned before, a target dataset with low $SI_0$ would be more challenging for a classifier compared to a target dataset with high $SI_0$.

- **Step2:** Considering all available pre-trained DNNs for a given target dataset, $\text{DNN}^j = \{\text{DNN}^1, \text{DNN}^2, \ldots, \text{DNN}^M\}$  $j = 1, 2, \ldots, M$

- **Step3:** Among $M$ available $\text{DNN}^j$, select randomly a new pre-trained $\text{DNN}^j$.

- **Step4:** Change size and input channel number of the target dataset according to the default of selected $\text{DNN}^j$.

- **Step5:** Insert the processed target dataset into selected $\text{DNN}^j$.

- **Step6:** Compute the separation index of target dataset ($SI_{N_L}^j$) in the last layer ($N_L$) of the selected $\text{DNN}^j$ and store it. The scheme of selected pre-trained DNNs (feature extracting part) is indicated in Fig.4.

- **Step7:** If all available pre-trained DNNs are chosen, go to the next step, otherwise go to step 3.

- **Step8:** Sort pre-trained DNNs in a descending order based on their $SI_{N_L}^j$ value

$$SI_{N_L}^{j_1^{**}} \geq SI_{N_L}^{j_2^{**}} \geq \ldots \geq SI_{N_L}^{j_T^{**}} \geq SI_0 \geq SI_{N_L}^{j_1^{*}} \geq SI_{N_L}^{j_2^{*}} \geq \cdots \geq SI_{N_L}^{j_N^{*}}$$

- **Step9:** Reject pre-trained DNNs whose $SI_{N_L}^j$ is less than the $SI_0$

$$SI_{N_L}^{j_N^*} \leq \cdots \leq SI_{N_L}^{j_2^*} \leq SI_{N_L}^{j_1^*} \leq SI_0$$

N: The number of rejected pre-trained DNNs

- **Step10:** Rank remained pre-trained DNNs based on the computed $SI_{N_L}^{j}$.

$$SI_{N_L}^{j_1^{**}} \geq SI_{N_L}^{j_2^{**}} \geq \ldots \geq SI_{N_L}^{j_T^{**}} \geq SI_0$$

T: The number of accepted pre-trained DNNs        T=M-N

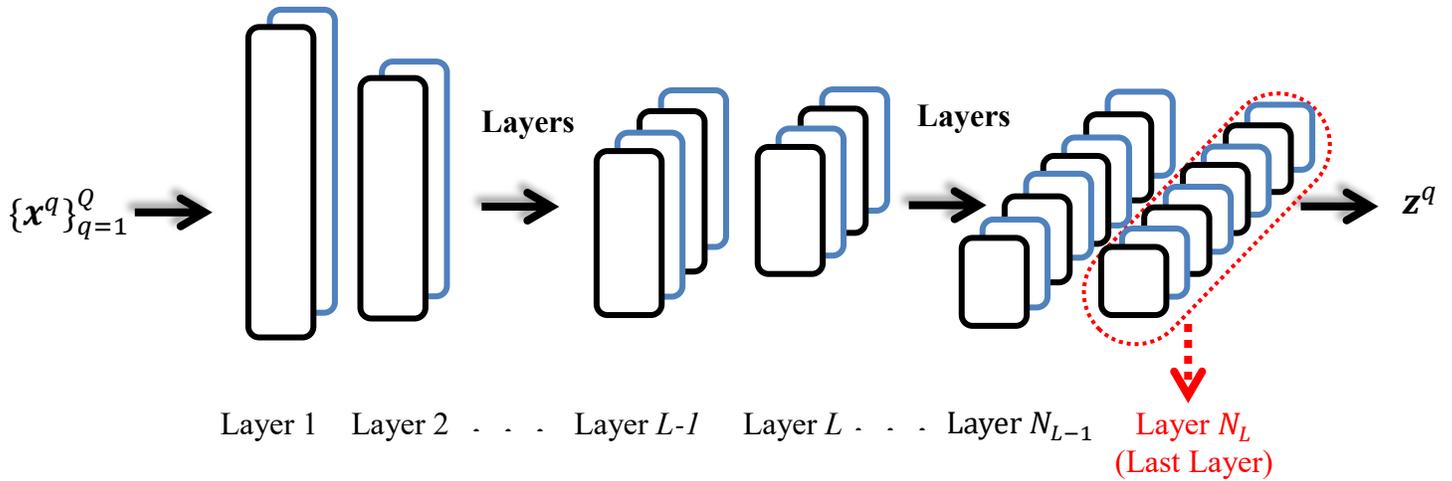

**Fig.4.** A scheme of selected a pre-trained $DNN^j$ (feature extracting part)

To investigate the efficiency of the proposed algorithm, the ranked pre-trained DNNs can be used for classification purposes. For this mean, a classifier model is designed to categorize a given target dataset as a number of labels. The diagram of a classifier model can be seen in Fig. 5. As it is understood from this figure, the classifier is a feed-forward deep neural network consisting of a feature extracting part (pre-trained DNNs) and a partitioning part. The classification process of this model can be explained in two parts.

In the first part, the target dataset ($\{x^q\}_{q=1}^{Q}$) is transformed to the feature extracting part (pre-trained DNN) in which the weights were learned on the source dataset. In this part, it is aimed to filter the distortions, nullities, and the disturbances of target dataset to achieve an effective feature space, which is indicated by $z^q$ in Fig. 5. In such conditions, the exclusive features of each label are amplified in comparison to common features with other labels. Consequently, the separation index of the target dataset can be increased layer by layer.

In the second part, the weights of the partitioning part are learned on the limited target dataset to make boundaries around the local regions including the dataset with the same labels. Finally, the resulted $\{\hat{l}^q\}_{q=1}^{Q}$ are compared with the target labels $\{l^q\}_{q=1}^{Q}$ to compute the classification accuracy of the classifier. In the next section, it will be seen that the results of the proposed algorithm match with the ranking of the classifiers by the classification accuracy on the target dataset.

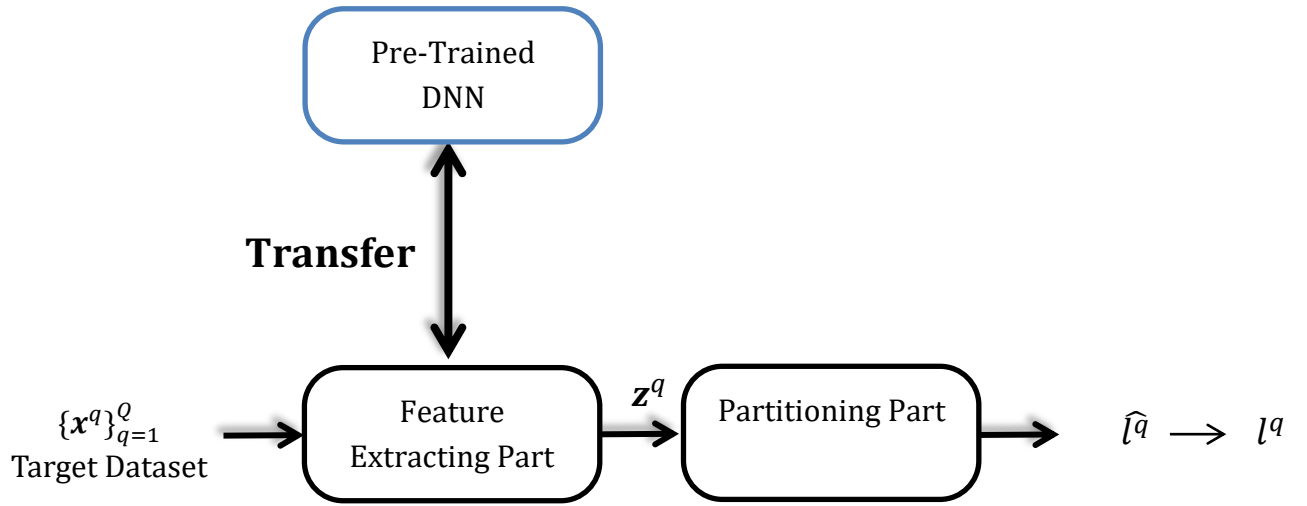

**Fig. 5.** A diagram of a model with two parts for classification purpose using transfer learning

## 3. Case studies

In this part, the main goal is to indicate the efficiency of the proposed algorithm in ranking and rejecting pre-trained DNNs for classification purpose. In addition, we are interested in investigating the effect of decreasing the number of target dataset on proposed algorithm. For this mean, three challenging target dataset including Linnaeus 5, Breast Cancer Images and COVID-CT Images are chosen. It is shown that there is a high correlation between the results of the proposed algorithm and classifiers performance after transfer learning.

*3.1: Target dataset - Linnaeus 5*

The first chosen target dataset is Linnaeus 5 (http://chaladze.com/l5/), which are divided into five major classes including berry, bird, dog, flower and other. For any classes, 1600 images are provided with the size of 256x256 pixels. A glimpse of the dataset is provided in Fig. 6.

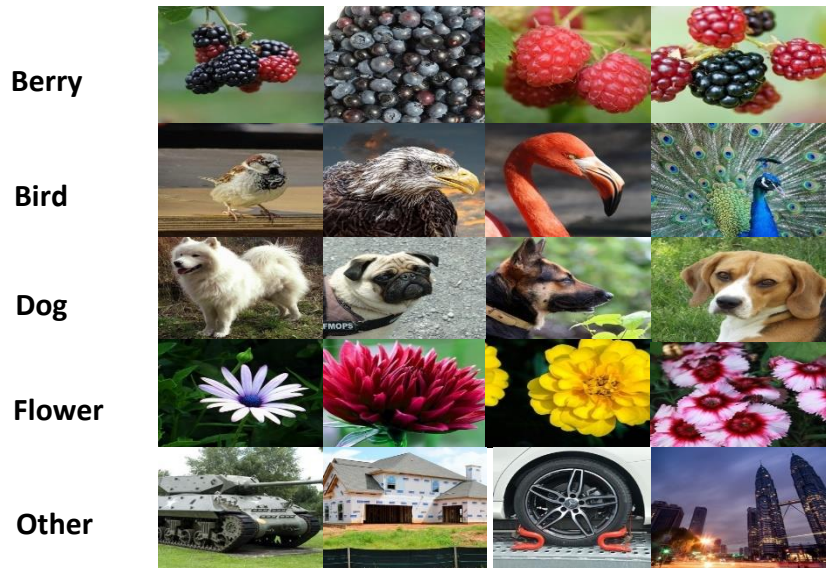

**Fig. 6.** Linnaeus 5 dataset with magnification level of 256x256

Here, In order to test the efficacy of the proposed algorithm, some popular DNNs trained over ImageNet such as VVG16, VGG19, InceptionV3, Xception, Resnet50 , DenseNet121 and EfficientNetB3 are considered . These pre-trained DNNs are suitable for transfer learning and readily available in packaged form through trusted public libraries such as Keras. According to the aforementioned algorithm, the separation index of Linnaeus 5 dataset in the first and final layers of considered pre-trained DNNs are computed and compared. Based on the calculated values, the proposed algorithm ranks the considered pre-trained DNNs in the order in which they probably impact the classification accuracy. Generally, it is expected that as $SI_{N_L}$ is increased, the classification accuracy of the classifiers is improved. From Table 1, it is clear that the EfficientB3 network is rejected due to reduction in the amount of separation index. In addition, it is predicted that the pre-trained Xception network outperforms the other networks.

Table 1. Comparing the classification performance achieved using the transfer learning with separation index of Linnaeus 5 dataset in the final layer of considered pre-trained DNNs.

| Pre-Trained DNNs | $SI_0$ | $SI_{N_L}$ | $SI_{N_L}$ (75% of dataset) | $SI_{N_L}$ (50% of dataset) | Classification Accuracy |
|---|---|---|---|---|---|
| **Xception** | | **0.94** | **0.930** | **0.924** | **97.2%** |
| InceptionV3 | | 0.92 | 0.897 | 0.881 | 96.9% |
| DenseNet121 | | 0.87 | 0.857 | 0.857 | 96.2% |
| VGG16 | **0.335** | 0.69 | 0.661 | 0.634 | 85.3% |
| VGG19 | | 0.68 | 0.652 | 0.625 | 83.6% |
| Resnet50 | | 0.39 | 0.372 | 0.363 | 42.3% |
| **EfficientB3** | | **0.32** | **0.316** | **0.313** | **18.6%** |

To show that the results of the proposed algorithm are reliable, a classifier model is designed containing considered pre-trained DNNs as a feature extracting part and partitioning part (see Fig.5). Fig.7 presents the general architecture of the classifier model. Before the learning process, the target dataset is split into training and validation dataset (20% = validation data, 80% = training data), which is usual practice for analysis of the performance of classifiers. As mentioned before, to compute the classification accuracy of classifiers models, only the parameters of partitioning part are learned over the training dataset. All the experiments are performed on the same machine (Google Colab) with configuration: Intel Xeon CPU clocked at 2.2 GHz, NVIDIA Tesla K80 GPU with 12GB GPU RAM, using Tensorflow and Keras library. Each classifier model is trained 50 epochs for effective analysis. Finally, after the learning process, the classification accuracy of classifiers models is computed. From Table.1, it can be said that the results of the proposed algorithm exactly match with the ranking of the trained DNNs by the accuracy on the target dataset.

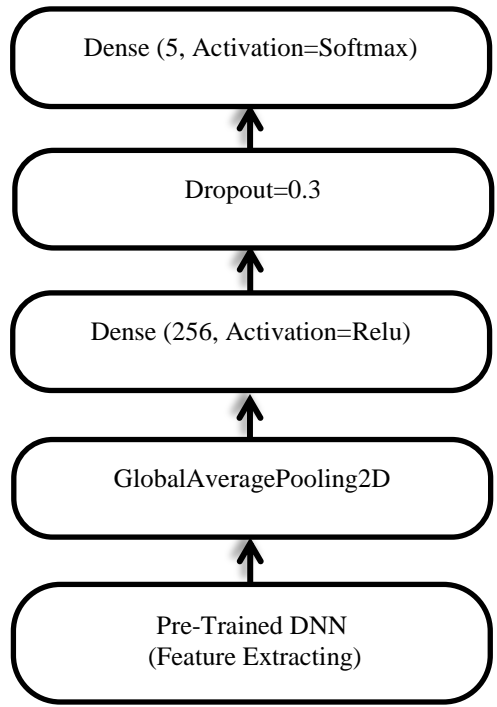

**Fig. 7.** The general architecture of classifier model

To study the performance of the proposed algorithm with respect to the number of target dataset, in two steps, 75% and 50% of the dataset is selected randomly to transform into the pre-trained DNNs. As it is clear from Table 1, with the reduction of target dataset numbers, the $SI_{N_L}$ of all considered pre-trained DNNs are decreased. However, it can be clearly observed that the proposed algorithm does not lose its efficiency in ranking pre-trained DNNs when a limited target dataset is available.

According to the proposed algorithm, the computed $SI_{N_L}$ and classification accuracy of pre-trained DNNs should obey a linear dependency. In other words, as the $SI_{N_L}$ is increased, the performance of networks is expected to improve. Fig.8. indicates clearly the high correlation between the classification accuracy and computed $SI_{N_L}$ as presented in Table 1.

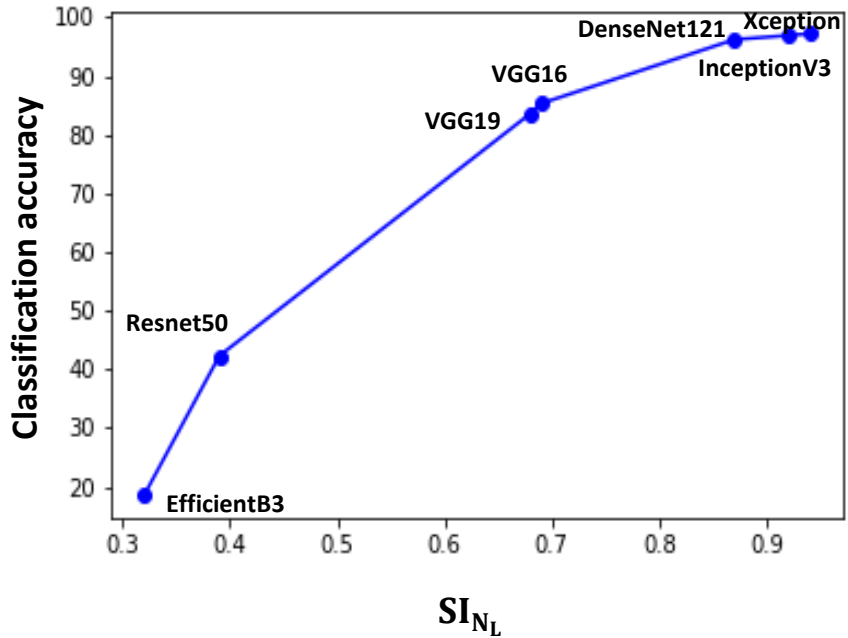

**Fig. 8.** The plot of separation index in the last layer of considered pre-trained DNNs versus their classification accuracy in test case 1

*3.2: Target dataset - Breast Cancer Images*

The procedure of data collection in the medical domain is more difficult compared to other domains because of restricted samples. So it is very challenging to access a well-annotated dataset in this domain. In this section, the selected target dataset is available from the BreakHis dataset (database-breakhis). All the 3 channel RGB images of breast cancer have a size of 700 × 460 and are divided into two groups including benign and malignant tumors which contain four different sub-datasets, namely 40, 100, 200, and 400X. A glimpse of the dataset can be seen in Fig.9.

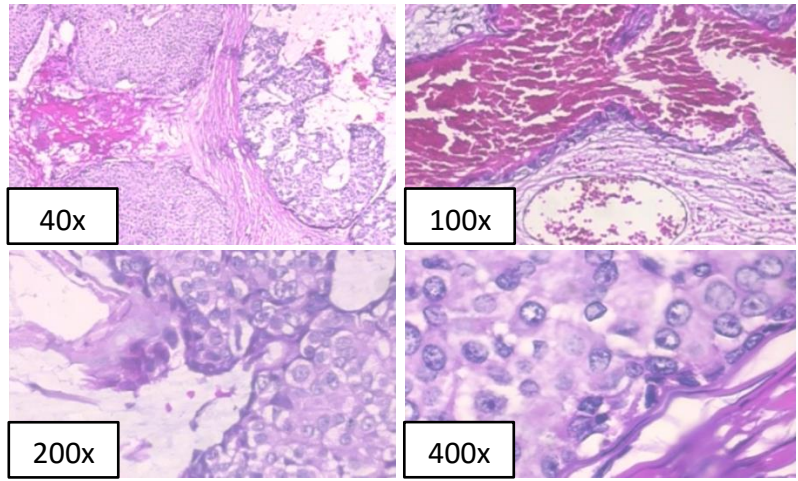

**Fig 9.** Breast Canser Histopathological Images from BreakHis Dataset of a patient suffered from malignant tumor with four magnification levels 40x, 100x, 200x and 400x

In [42], it is aimed to investigate which pre-trained DNNs have better performance in the classification of given challenging dataset. To fulfill this aim, three pre-trained DNNs including VGG16, VGG19 and ResNet50 are selected to classify the given target dataset. At first, for getting a balanced dataset, the number of image samples in the malignant class is decreased to the number of images in the benign class. All the procedure of transfer learning are carried out on the same machine (Intel(R) Core(TM) i7-7500U @ 2.90 GHz, NVIDIA GeForce 940MX, Window 10, 8 GB memory, using Tensorflow and Keras ). The details of the transfer learning process can be found in [42]. Finally, after a time-consuming procedure, it is revealed that the pre-trained VGG16 and ResNet50 are the best and the worst networks respectively to classify the given target dataset.

However, in our proposed algorithm, it is only enough to compute and compare the separation index of the target dataset in the end layers of selected pre-trained DNNs to rank them in a very short time. As presented in Table 2, the VGG16 network and the ResNet50 network have the most and the least amount of separation index in their final layers. The capability of the proposed algorithm can be seen from Fig.10 in which the x-axis reports the separation index in the last layer of pre-trained DNNs and the y-axis reports the computed classification accuracy. From this figure, it can be clearly found that the computed $SI_{N_L}$ exactly match with the classification accuracy. To demonstrate the insensitivity of the proposed algorithm to the number of target dataset, 75% and 50% of the dataset are chosen randomly to insert into the pre-trained DNNs .From Table 2, it can be seen that the proposed algorithm is able to rank the pre-trained DNNs with a limited target dataset.

**Table 2.** Comparing the classification performance achieved using the transfer learning with separation index of Breast Cancer Images dataset in the final layers of considered pre-trained DNNs.

| **Pre-trained DNNs** | $SI_0$ | $SI_{N_L}$ | $SI_{N_L}$ (75% of dataset) | $SI_{N_L}$ (50% of dataset) | **Classification Accuracy** |
|---|---|---|---|---|---|
| **VGG16** | | 0.719 | 0.692 | 0.677 | 92.6% |
| VGG19 | **0.682** | 0.700 | 0.681 | 0.672 | 90.0% |
| **ResNet50** | | 0.631 | 0.622 | 0.605 | 79.4% |

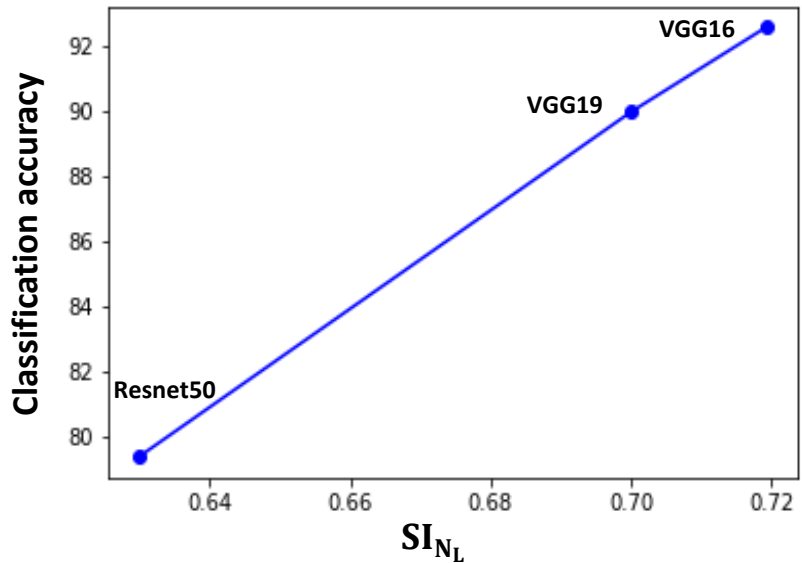

**Fig. 10.** The plot of separation index in the last layer of considered pre-trained DNNs versus their classification accuracy in test case 2

*3.3: Target dataset - COVID-CT Image*

The most challenging crisis that currently threatens human health is Coronavirus Disease (COVID19), which causes over 1.21M deaths around the world already (https://ourworldindata.org/covid-deaths). For this reason, it has been observed that a large number of researchers are motivated to work on this issue. In [44], some popular pre-trained DNNs are employed like VGG16, DenseNet121, VGG19, Xception, ResNet50v2, Inceptionv3 and NASNetLarge to investigate which of those can be better for classification of COVID-CT dataset. The target dataset consists of 397 negative and 346 positive cases, which is available in [43]. The size of images for both classes are various corresponding to height (maximum = 1853, average = 491, and minimum = 153) and width (maximum = 1485, average = 383, and minimum = 124). The illustrative example of the target dataset is presented in Fig.11. The details of data augmentation, the architecture of the partitioning part and the transfer learning process can be seen in [44]. After a time-consuming process, it becomes evident that the VGG16 network outperforms the other networks based on the F1 score.

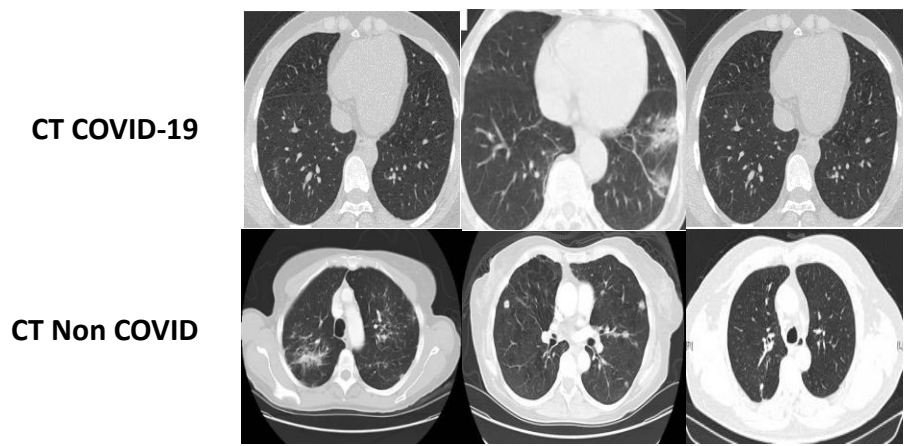

**Fig. 11.** The examples of CT scans that are positive and negative for COVID-19

However, these pre-trained DNNs can be ranked and rejected using the proposed concept in a short time without any training. From table 3, it is evident that the VGG16 network has the most separation index in comparison with other pre-trained networks . Besides, it is understood that the Resnet50V2 and NasnetLarge network are rejected for classifying the target dataset because of the reduction of the separation index. Generally, the results of Table 3 suggest that there is a high correlation between the ranking of the algorithm and the ranking resulted from classification accuracy, except in the Xception network. It is shown that while the $SI_{N_L}$ of Xception is more than $SI_{N_L}$ of the InceptionV3, InceptionV3 gives a better F1 score. The main reason for this issue would be that our algorithm are performed on the basic target dataset, while the accuracy of pre-trained DNNs is computed based on the dataset produced by the data augmentation technique. In general, this issue means that the more $SI_{N_L}$ of pre-trained DNNs are close to each other, the more challenging ranking of pre-trained DNNs can be.

**Table 3.** Comparing the classification performance achieved using the transfer learning with separation index of COVID-CT Images dataset in the final layers of considered pre-trained DNNs.

| Pre-trained DNNs | $SI_0$ | $SI_{N_L}$ | F1 score |
|---|---|---|---|
| **VGG16** | | 0.91 | 0.79 |
| VGG19 | | 0.88 | 0.78 |
| DenseNet121 | | 0.87 | 0.75 |
| InceptionV3 | **0.826** | 0.85 | 0.71 |
| **Xception** | | 0.87 | 0.70 |
| **Resnet50V2** | | 0.79 | 0.66 |
| **NasnetLarge** | | 0.75 | 0.64 |

## 4. Conclusion

In this paper, for ranking and rejecting pre-trained deep neural networks (DNNs) prior to utilizing them for a target dataset, a novel algorithm based on Separation Index (SI) was proposed. After giving a conceptual background about SI, the algorithm was explained. At first, the SI was computed for the last layer of feature extracted part of pre-trained DNNs and then they were ranked by descending sort on the computed SIs. According to this sorting, DNNs with higher SI were more qualified for transfer learning and a few DNNs with sufficiently low SIs may be rejected. The given algorithm was applied to three challenging datasets including Linnaeus 5, Breast Cancer Images and COVID-CT Images. The computed SIs for different pre-trained DNNs demonstrated that the resulted rankings from the algorithm match exactly with the accuracy ranking of the target data in the two first case studies. However, for the third case study, in spite of using different preprocessing on the target data, the resulted ranking had a high correlation with the accuracy ranking.